\documentclass[letterpaper, 10 pt, conference]{ieeeconf}  

\IEEEoverridecommandlockouts                              

\title{\name: Adversarial-Invariant Cross-Modal Alignment for Unified Robust Embeddings
}

\author{Yuhong Lu$^{1}$%
\thanks{$^{1}$Yuhong Lu is with Samueli School of Engineering, Electrical and Computer Engineering,
        UCLA, 7400 Boelter Hall, Los Angeles, CA 90024
        {\tt\small yuhong7ucla@g.ucla.edu}}%
}

\usepackage{array}
\usepackage{hyperref}
\usepackage{url}
\usepackage{tabularx}
\usepackage{threeparttable}
\usepackage{multirow}
\usepackage{makecell}
\usepackage{booktabs}
\usepackage{amsmath}
\usepackage{arydshln}
\usepackage{float}
\usepackage{pifont}
\usepackage{amsmath,amssymb}
\usepackage{colortbl}  
\usepackage{xcolor}

\usepackage{graphicx}
\usepackage{caption}

\newcommand{\name}{RLBind}

\begin{document}

\maketitle
\thispagestyle{empty}
\pagestyle{empty}

\begin{abstract}
Unified multi-modal encoders that bind vision, audio, and other sensors into a shared embedding space are attractive building blocks for robot perception and decision-making. However, on-robot deployment exposes the vision branch to adversarial and natural corruptions, making robustness a prerequisite for safety. Prior defenses typically align clean and adversarial features within CLIP-style encoders and overlook broader cross-modal correspondence, yielding modest gains and often degrading zero-shot transfer. We introduce \name, a two-stage adversarial-invariant cross-modal alignment framework for robust unified embeddings. Stage 1 performs unsupervised fine-tuning on clean–adversarial pairs to harden the visual encoder. Stage 2 leverages cross-modal correspondence by minimizing the discrepancy between clean/adversarial features and a text anchor, while enforcing class-wise distributional alignment across modalities. Extensive experiments on \textit{Image}, \textit{Audio}, \textit{Thermal}, and \textit{Video} data show that \name{} consistently outperforms the LanguageBind backbone and standard fine-tuning baselines in both clean accuracy and norm-bounded adversarial robustness. By improving resilience without sacrificing generalization, \name{} provides a practical path toward safer multi-sensor perception stacks for embodied robots in navigation, manipulation, and other autonomy settings.
\end{abstract}

\begin{figure*}[htbp]
\begin{center}
\includegraphics[width=0.95\textwidth]{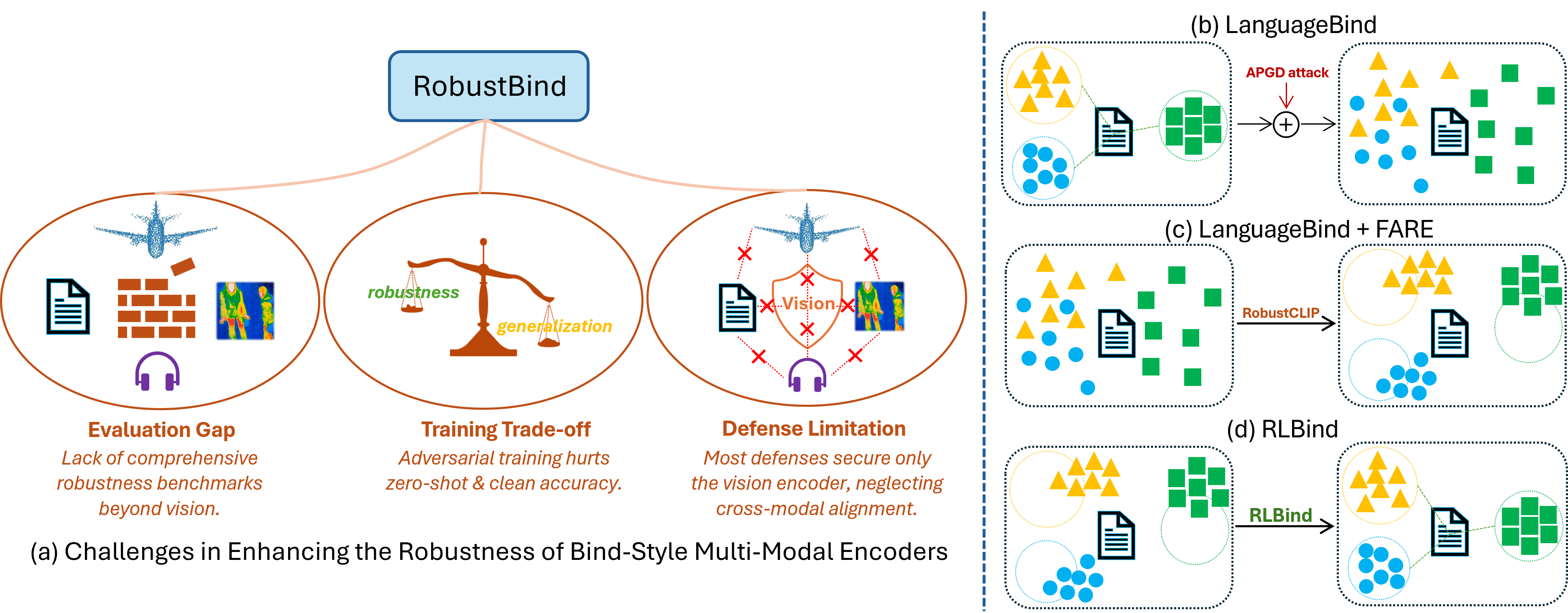}
\end{center}
\vspace{-8pt}
\caption{\textbf{(a)} Robustness in Bind-style multi-modal encoders is challenged by evaluation gaps, generalization-robustness trade-offs, and weak defenses. \textbf{(b)} Adversarial attacks disrupt LanguageBind's unified embedding space. \textbf{(c)} RobustCLIP improves robustness modestly but fails to recover LanguageBind's cross-modal correspondence, reducing zero-shot generalization. \textbf{(d)} \name{} leverages adversarial cross-modal alignment to preserve feature correspondence, achieving robustness gains without compromising generalization.}
\label{fig:concept}
\end{figure*}
\section{INTRODUCTION}

Multi-modal encoders have evolved significantly from the early CLIP~\cite{radford2021learning}, which unified vision and language, to ImageBind~\cite{girdhar2023imagebind}, which uses images as pivots, and more recently, LanguageBind~\cite{zhu2024languagebind}, which anchors all modalities to text. This text-centric approach has demonstrated strong multi-modal integration and superior benchmark performance. Despite their impressive capabilities, these large-scale multi-modal encoders remain highly vulnerable to adversarial perturbations, particularly on the vision side, raising concerns about the security of deploying such models in critical applications like robotics. Robotic systems, which rely on robust, real-time perception for tasks like navigation, manipulation, and human-robot interaction, face particular challenges in the presence of adversarial attacks on visual input. 

Research by~\cite{liao2025adversarial} shows that LanguageBind's accuracy on both images and videos collapses under adversarial attacks, with clean performance quickly degrading to near zero under moderate perturbation strengths. Similarly,~\cite{tsai2025enhance} argue that anchoring diverse modalities to text undermines robustness, especially under noise, temporal perturbations, and missing modalities. Consistent with these findings,~\cite{zhou2024revisiting} highlight that vision-language models (VLMs) are especially susceptible to vision-side adversarial attacks, underscoring the need for robust defenses such as MMCoA and RobustCLIP~\cite{schlarmann2024robustclip}.

In robotics, these vulnerabilities are particularly critical, as robots depend on accurate, cross-modal perception (vision, audio, and other sensors) to interact with dynamic and unpredictable environments. Despite advances, strengthening the robustness of Bind-style multi-modal encoders presents several challenges, as illustrated in Figure~\ref{fig:concept}(a). First, comprehensive robustness evaluations across diverse modalities (e.g., audio, thermal, point clouds) remain scarce, with systematic studies on LanguageBind still emerging. Second, adversarial training techniques often degrade zero-shot performance and risk overfitting, leading to a trade-off between robustness and generalization (e.g., TeCoA~\cite{mao2022understanding}). Third, most existing defenses focus solely on fortifying the vision encoder within a single modality (e.g., RobustCLIP~\cite{schlarmann2024robustclip}), neglecting the importance of cross-modal feature alignment, which limits their effectiveness in real-world robotic applications.

To address these challenges, we propose \name, a two-stage adversarial-invariant cross-modal alignment framework for learning unified robust embeddings within LanguageBind. In stage 1, we apply unsupervised fine-tuning on clean–adversarial embedding pairs, enhancing the model's robustness while minimizing zero-shot performance loss. In stage 2, we define cross-modal feature correspondence and establish adversarial-invariant distributional alignment to improve inter-modal robustness and restore generalization. 

\begin{figure*}[h!]
\begin{center}
\includegraphics[width=0.95\textwidth]{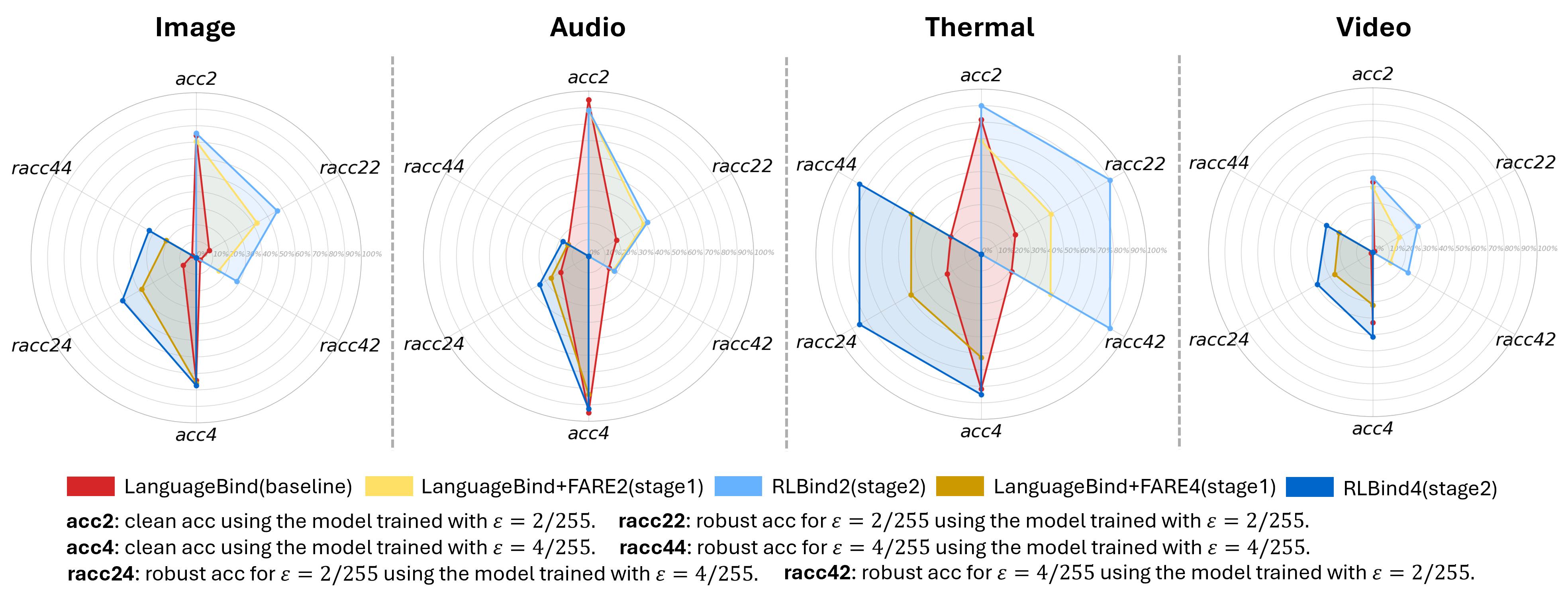}
\end{center}
\vspace{-8pt}
\caption{Comparison of our~\name(stage2), RobustCLIP~\cite{schlarmann2024robustclip}(stage1), and the baseline LanguageBind~\cite{zhu2024languagebind} across four modalities. \name{} achieves pronounced robustness improvements over both backbone model and unimodal fine-tuned baseline, while maintaining—or even enhancing—the clean-sample generalization of LanguageBind.}
\label{fig:radarmap}
\vspace{-12pt}
\end{figure*}

Extensive experiments across \textit{Image}, \textit{Audio}, \textit{Thermal}, and \textit{Video} data demonstrate that \name{} consistently outperforms LanguageBind and baseline fine-tuning approaches in both clean accuracy and norm-bounded adversarial robustness, as shown in Figure~\ref{fig:radarmap}. To the best of our knowledge, this is the \textbf{first work} to formally investigate the robustness of LanguageBind and successfully enhance it through cross-modal feature alignment, offering a critical step toward deploying robust multi-modal models in robotics.

\section{Related Work}

\subsection{Cross-modal Feature Alignment in Multi-modal LLMs: Bottlenecks and Advances with LanguageBind}

Large-scale cross-modal alignment was pioneered by contrastive dual-encoders such as CLIP~\cite{radford2021learning} and ALIGN~\cite{jia2021scaling}, which align image–text pairs at scale for zero-shot transfer. Yet, scaling these methods to LLMs with diverse modalities poses four major challenges. \textbf{i) Data Scarcity:} collecting aligned pairs for modalities like infrared, depth, and audio is costly, leading ImageBind~\cite{girdhar2023imagebind} to use images as a pivot, while major video corpora (e.g., HowTo100M~\cite{miech2019howto100m}, WebVid~\cite{bain2021frozen}, HD-VILA-100M~\cite{xue2022advancing}) rely on indirect textual signals such as ASR transcripts or web-scraped captions. \textbf{ii) Anchor Bias}: ImageBind~\cite{girdhar2023imagebind} relies on images as a pivot to align other modalities, which constrains cross-modal interactions and biases the shared space~\cite{lyu2024unibind}. \textbf{iii) Training Inefficiency:} fully training multi-modal LLMs is prohibitively expensive. BLIP-2~\cite{li2023blip} addresses this by freezing encoders and introducing a lightweight Q-Former, highlighting why inserting alignment losses directly into large models is non-trivial. \textbf{iv) Corrupted Modalities}: real deployments face missing, noisy, or asynchronous modalities; recent studies~\cite{wang2023multi}~\cite{10713849}~\cite{wu2024deep} show that models degrade without explicit training for missing-modality robustness. LangaugeBind~\cite{zhu2024languagebind} introduces a text-anchored alignment strategy that effectively alleviates the first three challenges. \textbf{i)}, LanguageBind is trained on the private VIDAL-10M dataset, which contains \textbf{10 million} aligned \textit{modality-language} pairs across five modalities (\textit{Video}, \textit{Infrared}, \textit{Depth}, \textit{Audio}, and \textit{Text}), thereby alleviating the bottleneck of non-visual aligned data. \textbf{ii)}, LanguageBind aligns all modalities directly to the language space, leveraging its rich semantic structure to improve cross-modal alignment and enhance downstream applicability. \textbf{iii)}, LanguageBind fixes the \textit{Text} encoder and applies LoRA for parameter-efficient fine-tuning of other modalities, reducing training inefficiency while preserving scalability for large multi-modal models. However, LanguageBind’s performance collapses under adversarial attacks, severely undermining its reliability and highlighting the urgent need to improve its robustness in adversarial settings.

\begin{figure*}[h!]
  \centering
  \includegraphics[width=\textwidth]{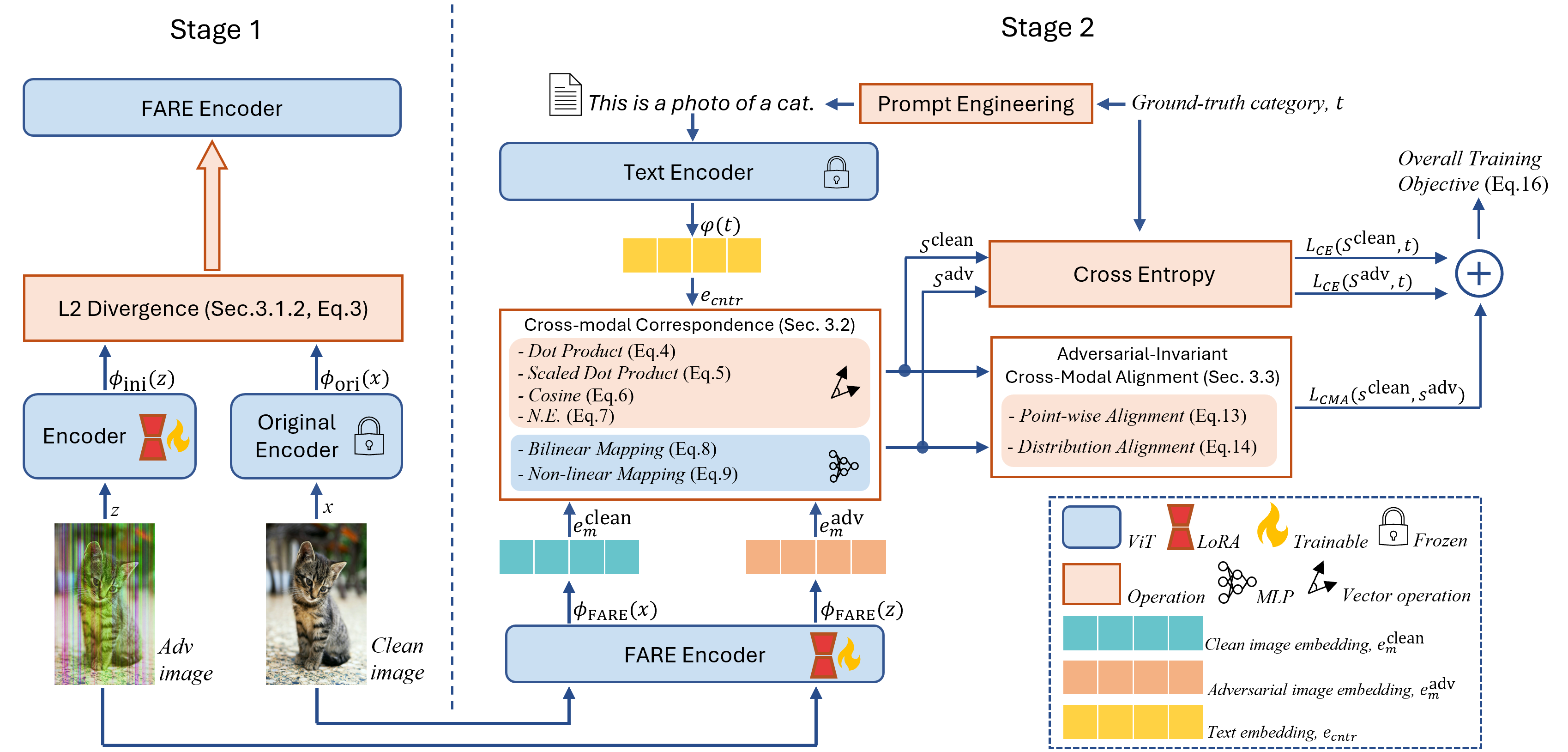}
  \caption{\name~pipeline for image-language alignment. Stage2 is initialized with the FARE encoder fine-tuned in stage1.}
  \vspace{-12pt}
  \label{fig:method}
\end{figure*}

\subsection{Adversarial Robustness in Multi-modal Encoders}

Prior efforts to improve the robustness of VLMs have mainly focused on the visual encoder. For instance,~\cite{mao2022understanding} proposed Text-guided Contrastive Adversarial Training(TeCoA), which generates adversarial image examples and aligns them with text via a contrastive loss. While it achieves notable gains in robust accuracy, its supervised, domain-specific fine-tuning comes at the expense of CLIP’s cross-domain generalization. For a better trade-off,~\cite{schlarmann2024robustclip} introduced FARE, an unsupervised adversarial fine-tuning method that minimizes the distance between clean and perturbed image embeddings. This produces a robust CLIP encoder that improves resistance to $L_{\infty}$ attacks while largely preserving zero-shot performance. Nevertheless, being confined to the visual encoder, FARE limits potential robustness gains by ignoring cross-modal alignment and leaves non-visual modalities vulnerable. The emergence of LanguageBind ~\cite{zhu2024languagebind} shifts this perspective. By leveraging VIDAL-10M and anchoring all modalities to the \textit{Text} space, LanguageBind builds a unified embedding space beyond vision–language pairs. This new setting invites us to rethink robustness not only as a property of the vision encoder but as a question of \textbf{cross-modal feature alignment}. In this work, we present RLBind, a two-stage adversarial-invariant cross-modal alignment framework that enhances LanguageBind’s robustness through clean–adversarial consistency while preserving generalization via cross-modal correspondence and distributional alignment. To the best of our knowledge, this is the \textbf{first work} to successfully improve the robustness of LanguageBind via adversarial-invariant cross-modal alignment. The most related effort is~\cite{liao2025adversarial}, which benchmark multiple models and propose an InfoNCE-based adversarial training scheme with LoRA to improve the robustness of UniBind~\cite{lyu2024unibind}.

\section{Method}

As illustrated in Figure~\ref{fig:method}, our \name{} is a two-stage adversarial-invariant cross-modal alignment framework designed for text-anchored multi-modal encoders. Taking \textit{image–text} alignment as an example, in stage 1, we adopt FARE’s unsupervised fine-tuning~\cite{schlarmann2024robustclip}(Sec.~\ref{sec: RobustCLIP}) on clean–adversarial(APGD,~\cite{croce2020reliable}) pairs, yielding a robustness-enhanced FARE encoder. Passing a clean sample and its APGD perturbation through the FARE encoder yields their respective embeddings. Stage 2 further fine-tunes this encoder by leveraging the discrepancy between the clean and adversarial embeddings in a supervised way. Specifically, we compute the similarities of the clean and adversarial embeddings to the text anchor via cross-modal correspondence(Sec.~\ref{sec: Cross-modal Correspondence Definition}), then enforce class-wise distributional alignment with adversarial-invariant cross-modal alignment(Sec.~\ref{sec: Adversarial‑Invariant Cross‑Modal Alignment}). Together with cross-entropy on clean and adversarial predictions, this alignment term constitutes the overall training objective in stage 2.

\subsection{Preliminary}

\subsubsection{Contrastive Learning in LanguageBind}

LanguageBind~\cite{zhu2024languagebind} aligns embeddings of \textit{Video}, \textit{Infrared}, \textit{Depth}, and \textit{Audio} with their corresponding \textit{Language} through the contrastive learning InfoNCE loss as
\begin{align}
L_{M2T} = -\frac{1}{K}\sum_{i=1}^K
   \log\frac{\exp\bigl(x_i^\top y_i / \tau\bigr)}
            {\sum_{j=1}^K \exp\bigl(x_i^\top y_j / \tau\bigr)}, 
\; \\
L_{T2M} = -\frac{1}{K}\sum_{i=1}^K
   \log\frac{\exp\bigl(y_i^\top x_i / \tau\bigr)}
            {\sum_{j=1}^K \exp\bigl(y_i^\top x_j / \tau\bigr)} \notag
\label{eq:LanguageBind-infoNCE}
\end{align}
where $x_{i}$ and $y_{j}$ denote the \textit{i}-th modality data and the \textit{j}-th text with their features normalized, while $K$ is the batch size and $\tau$ the temperature. 
LanguageBind takes \textit{Language} as the semantic center and jointly fine-tunes encoders of other modalities to align their embeddings with the unified language-centered embedding space. Trained on the private VIDAL-10M with 10M multimodal text pairs, LanguageBind achieves superior performance across 15 benchmarks.

\subsubsection{RobustCLIP with FARE\label{sec: RobustCLIP}}

RobustCLIP~\cite{schlarmann2024robustclip} enhances CLIP encoders’ adversarial robustness through Fine-tuning for Adversarially Robust Embeddings (FARE). It reframes the supervised objective of matching adversarial and original image–text cosine-similarities into an unsupervised problem of minimizing the $L_{2}$ distance between perturbed and original image embeddings, supported by
\begin{align}
\bigl|\cos\bigl(\phi_{\mathrm{FT}}(z), \psi(t)\bigr) - \cos\bigl(\phi_{\mathrm{Org}}(x), \psi(t)\bigr)\bigr|
\le \\
\min\Bigl(\tfrac{2}{\|\phi_{\mathrm{Org}}(x)\|_{2}},\,\tfrac{2}{\|\phi_{\mathrm{FT}}(z)\|_{2}}\Bigr)\,\bigl\|\phi_{\mathrm{FT}}(z) - \phi_{\mathrm{Org}}(x)\bigr\|_{2}\, \notag
\label{eq: RobustCLIP principle}
\end{align}
where $\phi_{\mathrm{FT}}$ denotes the vision encoder to be fine-tuned, $\phi_{\mathrm{Org}}$ the original frozen OpenCLIP vision encoder, and $\psi$ the fixed text encoder. Therefore, given an image $x$, FARE's training objective goes without the supervised text label $t$ and is reduced to
\begin{equation}
L_{\mathrm{FARE}}(\phi, x) = \max_{\|z - x\|_{\infty} \le \varepsilon} \bigl\|\phi(z) - \phi_{\mathrm{Org}}(x)\bigr\|_{2}^{2}
\label{eq:RobustCLIP LossFunc}
\end{equation}
where $\varepsilon$ is the perturbation size. As an unsupervised approach, FARE is not bound by ImageNet labels and therefore generalizes better than supervised methods like TeCoA~\cite{mao2022understanding}.

\subsection{Cross-modal Correspondence Definition}\label{sec: Cross-modal Correspondence Definition}
Suppose $e_1 \in \mathbb{R}^d$ is an embedding of $\text{Modality}_1$ and $e_2 \in \mathbb{R}^d$ an embedding of $\text{Modality}_2$. They share an common embedding dimension $d$. The cross-modal correspondence score can be calculated through one of the following 6 ways.

\noindent\textbf{(a) Dot-Product}
\begin{equation}
s_{\mathrm{dot}}(e_1,e_2) = e_1^\top e_2.
\label{eq:dot product}
\end{equation}
This simple inner product has no trainable parameters and is highly efficient, but its raw scale may vary with embedding norms.

\noindent\textbf{(b) Scaled Dot‑Product}
\begin{equation}
s_{\mathrm{scaled-dot}}(e_1,e_2;\alpha) = \alpha \cdot e_1^\top e_2,
\label{eq:scaled dot product}
\end{equation}
where $\alpha = 1/\sqrt{d}$ or learned. The normalization reduces variance in magnitude, improving numerical stability.

\noindent\textbf{(c) Cosine Closeness}
\begin{equation}
s_{\mathrm{cos}}(e_1,e_2) 
  = \frac{e_1^\top e_2}{\|e_1\|_2\,\|e_2\|_2} \in [-1, 1].
\label{eq:cosine closeness}
\end{equation}
By explicitly normalizing both embeddings to $[0, 1]$ before comparing, this measurement is completely irrelevant to embedding magnitudes.

\noindent\textbf{(d) Normalized Euclidean(NE) Distance}
\begin{equation}
s_{\mathrm{euclid}}(e_1, e_2) = 1 - \frac{\|\,e_1 - e_2\|_2}{\max\bigl(\|e_1\|_2,\|e_2\|_2\bigr)} \in [0, 1].
\label{eq:euclid distance}
\end{equation}
Both (c) and (d) yield bounded similarity scores. While Cosine Closeness measures only the angle between features, NE reflects their actual spatial separation, penalizing differences in length as well as direction.

\noindent\textbf{(e) Bilinear Mapping}
\begin{equation}
s_{\mathrm{bilinear}}(e_1,e_2; W) = e_1^\top W\,e_2
\label{eq:bilinear}
\end{equation}
with $W\in\mathbb{R}^{d\times d}$ a learned weight matrix. It linearly transforms one feature before computing the correspondence, improving expressiveness at the expense of additional $O(d^2)$ parameters.

\noindent\textbf{(f) Non-linear Mapping}
\begin{equation}
s_{\mathrm{MLP}}(e_1,e_2; \Theta) = \mathrm{MLP}\bigl([e_1: e_2]; \Theta \bigr)
\label{eq:mlp}
\end{equation}
where $[\,e_1:e_2\,]\in\mathbb{R}^{2d}$ denotes the concatenation of two modality features, and $\Theta$ represents all parameters of the MLP. While MLPs capture complex non-linear relations, they incur higher computation and need regularization. 

Based on the definitions of these six correspondence methods, we subsequently develop an adversarial-invariant cross-modal alignment framework for robustness fine-tuning.
\begin{table*}[ht!]
\centering
\caption{RLBind’s performance across four modalities under attacks of varying strength.}
\scriptsize
\setlength{\tabcolsep}{16pt}
\resizebox{\linewidth}{!}{%
\begin{tabular}{l l l l l}
\toprule
\textbf{Method} & \textbf{Modality} & \textbf{Clean Acc.} 
& \makecell{\textbf{Robust Acc.} \\ $\varepsilon_1 = \tfrac{2}{255}$} 
& \makecell{\textbf{Robust Acc.} \\ $\varepsilon_2 = \tfrac{4}{255}$} \\
\midrule

\multirow{4}{*}{LanguageBind (baseline)} 
& Image   & 74.07\% & 9.12\% & 2.84\% \\
& Audio   & 94.75\% & 19.45\% & 14.25\% \\
& Thermal & 81.66\% & 23.86\% & 21.22\% \\
& Video   & 42.74\% & 1.21\%  & 0.13\% \\
\midrule

\multirow{4}{*}{LanguageBind + FARE2 (stage1)$^{1}$} 
& Image   & 70.57\%(↓3.50) & 42.49\%(↑33.37) & 15.72\%(↑12.88) \\
& Audio   & 88.25\%(↓6.50) & 38.50\%(↑19.05) & 17.25\%(↑3.00) \\
& Thermal & 68.55\%(↓13.11)& 48.68\%(↑24.82) & 48.43\%(↑27.21) \\
& Video   & 39.60\%(↓3.14) & 18.49\%(↑17.28) & 12.43\%(↑12.30) \\
\hdashline
\multirow{4}{*}{LanguageBind + FARE4 (stage1)$^{2}$} 
& Image   & 75.98\%(↑1.91) & 38.14\%(↑29.02) & 21.07\%(↑18.23) \\
& Audio   & 83.75\%(↓11.00)& 26.25\%(↑6.80)  & 14.50\%(↑0.25)  \\
& Thermal & 62.52\%(↓19.14)& 48.97\%(↑25.11) & 48.81\%(↑27.59) \\
& Video   & 32.10\%(↓10.64)& 26.81\%(↑25.60) & 23.83\%(↑23.70) \\
\midrule

\multirow{4}{*}{RLBind2 (stage2)$^{3}$} 
& Image   & \textbf{75.43\%(↑1.36)} & \textbf{56.76\%(↑47.64)} & \textbf{28.49\%(↑25.65)} \\
& Audio   & \textbf{88.50\%(↓6.25)} & \textbf{41.25\%(↑21.80)} & \textbf{18.00\%(↑3.75)}  \\
& Thermal & \textbf{90.00\%(↑8.34)} & \textbf{90.00\%(↑66.14)} & \textbf{90.00\%(↑68.78)} \\
& Video   & \textbf{45.26\%(↑2.52)} & \textbf{31.58\%(↑30.37)} & \textbf{24.74\%(↑24.61)} \\
\hdashline
\multirow{4}{*}{RLBind4 (stage2)$^{4}$} 
& Image   & \textbf{77.39\%(↑3.32)} & \textbf{51.62\%(↑42.50)} & \textbf{33.02\%(↑30.18)} \\
& Audio   & \textbf{92.50\%(↓2.25)} & \textbf{34.25\%(↑14.80)} & \textbf{18.00\%(↑3.75)}  \\
& Thermal & \textbf{85.00\%(↑3.34)}  & \textbf{85.00\%(↑61.14)} & \textbf{85.00\%(↑63.78)} \\
& Video   & \textbf{51.58\%(↑8.84)}  & \textbf{38.95\%(↑37.74)} & \textbf{32.63\%(↑32.50)} \\
\bottomrule
\end{tabular}%
}

\vspace{0.5em}
\raggedright
{\scriptsize
$^{1}$ LanguageBind + FARE2(stage1): fine-tuning with FARE under APGD attack of radii $\varepsilon = 2/255$ as stage1. \\
$^{2}$ LanguageBind + FARE4(stage1): fine-tuning with FARE under APGD attack of radii $\varepsilon = 4/255$ as stage1. \\
$^{3}$ RLBind2(stage2): stage2 fine-tuning, initialized with LanguageBind + FARE2(stage1) under APGD attack of radii $\varepsilon = 2/255$. \\
$^{4}$ RLBind4(stage2): stage2 fine-tuning, initialized with LanguageBind + FARE4(stage1) under APGD attack of radii $\varepsilon = 4/255$. \\
}
\label{tab:rlbind_perf}
\vspace{-12pt}
\end{table*}

\subsection{Adversarial‑Invariant Cross‑Modal Alignment}\label{sec: Adversarial‑Invariant Cross‑Modal Alignment}

Assume $e_m^{\text{clean}}, e_m^{\text{adv}}\in \mathbb{R}^d$ are two embeddings of a sample before and after the norm-bounded adversarial perturbation in the $\text{modality}_m$ that is to be aligned with the anchor modality. $e_{\text{cntr},i} \in \mathbb{R}^d$ is the $i^{\text{th}}$ embedding in the anchor modality, describing the semantics of the $i^{\text{th}}$ class. Concatenate anchor embeddings of all $C$ classes yields the anchor matrix, $E_{\text{cntr}}\in \mathbb{R}^{d \times C}$ that collects the semantics of all $C$ classes. We enforce
\begin{equation}
s(e_m^{\text{clean}}, E_{\text{cntr}}[:, c]) \approx s(e_m^{\text{adv}}, E_{\text{cntr}}[:, c])
\label{eq:adversarial-invariant cross-modal alignment_1}    
\end{equation}
holds for all $c \in C$, where $s$ is the cross-modal correspondence score defined as one of the Eq.~\ref{eq:dot product} to Eq.~\ref{eq:mlp} in Sec.~\ref{sec: Cross-modal Correspondence Definition}. In other words, we enforce that clean and adversarial embeddings yield similar cross-modal correspondence scores against the anchor within the whole distribution range of classes. Denote
\begin{equation}\label{eq:adversarial-invariant cross-modal alignment_2}
s(e_m^{\text{clean}}, E_{\text{cntr}}[:, c]) := s^{\text{clean},c}, \;
s(e_m^{\text{adv}}, E_{\text{cntr}}[:, c]) := s^{\text{adv},c}
\end{equation}
and concisely, for $C$ classes,
\begin{align}\label{eq:adversarial-invariant cross-modal alignment_3}
s^{\mathrm{clean}} = \bigl[s^{\mathrm{clean},1},\,\dots,\,s^{\mathrm{clean},C}\bigr], \; \\ 
s^{\mathrm{adv}} = \bigl[s^{\mathrm{adv},1},\,\dots,\,s^{\mathrm{adv},C}\bigr]. \notag
\end{align}
Then the alignment can be formulated in one of the following two methods.

\noindent\textbf{(a) Point-wise Level Alignment}
\begin{equation}\label{eq:point-wise distribution alignment}
\mathcal{L}_{p}
\;=\;
\frac{1}{C}
\sum_{i=1}^{C}
\bigl|s^{\mathrm{clean},i} \;-\; s^{\mathrm{adv},i}\bigr|^{p}
\quad\text{for }p\in\{1,2\}.
\end{equation}
By selecting $p=1\;\text{or}\;2$, $\mathcal{L}_{p}$ enforces point-wise alignment while offering a tunable robustness–sensitivity trade-off.

\noindent\textbf{(b) Distribution Level Alignment}

Instead of point-wise $L_{p}$ alignment, we can alternatively align score distributions via KL divergence. Raw scores are normalized:
\begin{align}\label{eq:kl_alignment_1}
P_{i} \;=\;\frac{\exp\bigl(s^{\mathrm{clean},i}/\tau'\bigr)}%
{\sum_{j=1}^{C}\exp\bigl(s^{\mathrm{clean},j}/\tau'\bigr)},\\ \quad
Q_{i} \;=\;\frac{\exp\bigl(s^{\mathrm{adv},i}/\tau'\bigr)}%
{\sum_{j=1}^{C}\exp\bigl(s^{\mathrm{adv},j}/\tau'\bigr)}, \notag
\end{align}
where $\tau'$ is a temperature. The symmetric KL loss is
\begin{align}\label{eq:kl_alignment_2}
\mathcal{L}_{\mathrm{KL}}
\;=\;
\mathrm{KL}(P \,\|\,Q) \;+\;\mathrm{KL}(Q \,\|\,P) \\ 
\;=\;
\sum_{i=1}^{C}P_{i}\log\frac{P_{i}}{Q_{i}}
\;+\;\sum_{i=1}^{C}Q_{i}\log\frac{Q_{i}}{P_{i}}\,. \notag
\end{align}
Minimizing \(\mathcal{L}_{\mathrm{KL}}\) aligns clean and adversarial distributions, discouraging both under- and over-confidence.

\begin{table*}[htbp]
\centering
\setlength{\tabcolsep}{14pt}
\caption{Performance comparison showing consistent accuracy gains of RLBind(stage2) over LanguageBind + FARE(stage1) across modalities.}
\scriptsize
\resizebox{\linewidth}{!}{%
\begin{tabular}{l l l l l}
\toprule
\textbf{Method} & \textbf{Modality} & \textbf{Clean Acc.} 
& \makecell{\textbf{Robust Acc.} \\ $\varepsilon_1 = \tfrac{2}{255}$} 
& \makecell{\textbf{Robust Acc.} \\ $\varepsilon_2 = \tfrac{4}{255}$} \\
\midrule

\multirow{4}{*}{LanguageBind + FARE2 (Stage1)} 
& Image   & 70.57\% & 42.49\% & 15.72\% \\
& Audio   & 88.25\% & 38.50\% & 17.25\% \\
& Thermal & 68.55\% & 48.68\% & 48.43\% \\
& Video   & 39.60\% & 18.49\% & 12.43\% \\
\hdashline
\multirow{4}{*}{RLBind2 (Stage2)} 
& Image   & \textbf{75.43\%(↑4.86)} & \textbf{56.76\%(↑14.27)} & \textbf{28.49\%(↑12.77)} \\
& Audio   & \textbf{88.50\%(↑0.25)} & \textbf{41.25\%(↑2.75)}  & \textbf{18.00\%(↑0.75)}  \\
& Thermal & \textbf{90.00\%(↑21.45)}& \textbf{90.00\%(↑41.32)} & \textbf{90.00\%(↑41.57)} \\
& Video   & \textbf{45.26\%(↑5.66)} & \textbf{31.58\%(↑13.09)} & \textbf{24.74\%(↑12.31)} \\

\midrule

\multirow{4}{*}{LanguageBind + FARE4 (Stage1)} 
& Image   & 75.98\% & 38.14\% & 21.07\% \\
& Audio   & 83.75\% & 26.25\% & 14.50\% \\
& Thermal & 62.52\% & 48.97\% & 48.81\% \\
& Video   & 32.10\% & 26.81\% & 23.83\% \\
\hdashline
\multirow{4}{*}{RLBind4 (Stage2)} 
& Image   & \textbf{77.39\%(↑1.41)} & \textbf{51.62\%(↑13.48)} & \textbf{33.02\%(↑11.95)} \\
& Audio   & \textbf{92.50\%(↑8.75)} & \textbf{34.25\%(↑8.00)}  & \textbf{18.00\%(↑3.50)}  \\
& Thermal & \textbf{85.00\%(↑22.48)}& \textbf{85.00\%(↑36.03)} & \textbf{85.00\%(↑36.19)} \\
& Video   & \textbf{51.58\%(↑19.48)}& \textbf{38.95\%(↑12.14)} & \textbf{32.63\%(↑8.80)} \\
\bottomrule
\end{tabular}%
}
\label{tab:abl_stg}
\end{table*}

\begin{table*}[h!]
\centering
\caption{\name{} performance across cross-modal correspondence and adversarial-invariant alignment variants.}
\scriptsize
\resizebox{\linewidth}{!}{%
\begin{tabular}{l l l c c c}
\toprule
\textbf{Method} & \textbf{Cross-modal Correspondence} & \makecell{\textbf{Adversarial-Invariant}\\\textbf{Cross-modal Alignment}}
& \textbf{Clean Acc.} 
& \makecell{\textbf{Robust Acc.}\\$\varepsilon_1 = \tfrac{2}{255}$}
& \makecell{\textbf{Robust Acc.}\\$\varepsilon_2 = \tfrac{4}{255}$} \\
\midrule

\multirow{18}{*}{RLBind2 (Stage2)} 
& \multirow{3}{*}{Dot-Product} & L1 & \textbf{75.83\%} & 56.58\% & 27.85\% \\
& & L2 & 75.43\% & 56.76\% & 28.49\% \\
& & KL divergence & 74.27\% & \textbf{57.58\%} & \textbf{32.07\%} \\
\cdashline{2-6}
& \multirow{3}{*}{Scaled Dot-Product} & L1 & 66.44\% & 43.84\% & 14.61\% \\
& & L2 & 64.86\% & 42.33\% & 13.89\% \\
& & KL divergence & 66.14\% & 45.36\% & 16.23\% \\
\cdashline{2-6}
& \multirow{3}{*}{Cosine} & L1 & 66.01\% & 44.33\% & 19.50\% \\
& & L2 & 65.66\% & 43.69\% & 18.50\% \\
& & KL divergence & 66.23\% & 44.94\% & 20.03\% \\
\cdashline{2-6}
& \multirow{3}{*}{Normalized Euclidean} & L1 & 61.21\% & 39.75\% & 18.38\% \\
& & L2 & 60.60\% & 39.48\% & 18.31\% \\
& & KL divergence & 61.15\% & 40.01\% & 19.03\% \\
\cdashline{2-6}
& \multirow{3}{*}{Bilinear Mapping} & L1 & 71.41\% & 50.60\% & 22.17\% \\
& & L2 & 70.56\% & 51.43\% & 23.68\% \\
& & KL divergence & 69.91\% & 51.27\% & 24.66\% \\
\cdashline{2-6}
& \multirow{3}{*}{Non-linear Mapping} & L1 & 24.68\% & 16.82\% & 10.46\% \\
& & L2 & 26.88\% & 19.15\% & 13.80\% \\
& & KL divergence & 51.22\% & 36.14\% & 19.29\% \\

\midrule

\multirow{18}{*}{RLBind4 (Stage2)} 
& \multirow{3}{*}{Dot-Product} & L1 & \textbf{79.08\%} & 51.60\% & 33.23\% \\
& & L2 & 77.39\% & 51.62\% & 33.02\% \\
& & KL divergence & 69.88\% & \textbf{55.07\%} & \textbf{38.44\%} \\
\cdashline{2-6}
& \multirow{3}{*}{Scaled Dot-Product} & L1 & 69.68\% & 40.23\% & 23.18\% \\
& & L2 & 69.66\% & 40.64\% & 23.28\% \\
& & KL divergence & 70.19\% & 42.02\% & 24.45\% \\
\cdashline{2-6}
& \multirow{3}{*}{Cosine} & L1 & 68.95\% & 39.74\% & 23.40\% \\
& & L2 & 69.04\% & 39.80\% & 23.47\% \\
& & KL divergence & 70.01\% & 40.72\% & 23.96\% \\
\cdashline{2-6}
& \multirow{3}{*}{Normalized Euclidean} & L1 & 64.08\% & 37.57\% & 21.61\% \\
& & L2 & 63.95\% & 37.68\% & 21.84\% \\
& & KL divergence & 64.10\% & 38.00\% & 22.05\% \\
\cdashline{2-6}
& \multirow{3}{*}{Bilinear Mapping} & L1 & 74.81\% & 41.58\% & 24.62\% \\
& & L2 & 68.91\% & 42.85\% & 25.30\% \\
& & KL divergence & 64.50\% & 44.70\% & 28.59\% \\
\cdashline{2-6}
& \multirow{3}{*}{Non-linear Mapping} & L1 & 11.60\% & 5.46\% & 3.56\% \\
& & L2 & 11.62\% & 7.23\% & 5.23\% \\
& & KL divergence & 39.86\% & 22.21\% & 13.50\% \\
\bottomrule
\end{tabular}%
}
\label{tab:abl_aca}
\end{table*}

\begin{table*}[h!]
\centering
\setlength{\tabcolsep}{10pt}
\caption{Comparison of RLBind’s performance across different training objective configurations.}
\scriptsize
\resizebox{\linewidth}{!}{%
\begin{tabular}{l c c c c c c}
\toprule
\textbf{Method} & \textbf{clean c.e.} & \textbf{adv c.e.} & \textbf{CMA} 
& \textbf{Clean Acc.} 
& \makecell{\textbf{Robust Acc.}\\$\varepsilon_1 = \tfrac{2}{255}$}
& \makecell{\textbf{Robust Acc.}\\$\varepsilon_2 = \tfrac{4}{255}$} \\
\midrule

\multirow{3}{*}{RLBind2 (Stage2)} 
& \checkmark &            &            & \textbf{79.84\%} & 26.66\% & 13.12\% \\
& \checkmark & \checkmark &            & 74.80\% & 56.12\% & 27.57\% \\
& \checkmark & \checkmark & \checkmark & 75.43\% & \textbf{56.76\%} & \textbf{28.49\%} \\
\midrule

\multirow{3}{*}{RLBind4 (Stage2)} 
& \checkmark &            &            & \textbf{80.20\%} & 40.20\% & 14.97\% \\
& \checkmark & \checkmark &            & 76.86\% & 49.48\% & 32.58\% \\
& \checkmark & \checkmark & \checkmark & 77.39\% & \textbf{51.62\%} & \textbf{33.02\%} \\
\bottomrule
\end{tabular}%
}

\vspace{0.5em}
\raggedright
{\scriptsize
* Experiments are with Dot-Product (Eq.\ref{eq:dot product}) as Cross-modal Correspondence and L2 (Eq.\ref{eq:kl_alignment_2}, $p=2$) as Adversarial-Invariant Cross-modal Alignment. \\
}
\label{tab:abl_lossfunc}
\vspace{-12pt}
\end{table*}

\subsection{Overall Training Objective}

In stage1, Eq.\ref{eq:RobustCLIP LossFunc} is applied to all non-anchor modality encoders of LanguageBind~\cite{zhu2024languagebind} to perform unsupervised fine-tuning on clean–adversarial embedding pairs, enabling LanguageBind to gain robustness through FARE~\cite{schlarmann2024robustclip} with while incurring minor zero-shot degradation. Upon convergence, a robustness-enhanced multi-modal encoder is obtained. In stage2, clean and adversarial samples are passed through this encoder, and their embeddings are refined via adversarial-invariant cross-modal alignment, optimized using the following loss function:
\begin{align}\label{eq: lossfunc for 2nd}
\mathcal{L}_{2^{\text{nd}}\;\text{stage}}(e_m^{\text{clean}}, e_m^{\text{adv}}, e_{\text{cntr}}, t) = \mathcal{L}_{\text{CE}}(e_m^{\text{clean}}, t) + \mathcal{L}_{\text{CE}}(e_m^{\text{adv}}, t) \\ +\lambda\cdot\mathcal{L}_{\text{CMA}}(e_m^{\text{clean}}, e_m^{\text{adv}}, e_{\text{cntr}}) \notag
\end{align}
where $e_m^{\text{clean}}$, $e_m^{\text{adv}}$ are clean and adversarial embeddings for a sample in the $\text{modality}_m$ to be aligned with $e_{\text{cntr}}$, its embedding in the anchor modality. $\mathcal{L}_{\text{CE}}$ is the cross-entropy loss quantifying the divergence between the predicted probability distribution and the target $t$. $\mathcal{L}_{\text{CMA}}$ denotes the loss function for adversarial-invariant cross-modal alignment, which is specified as either Eq.\ref{eq:point-wise distribution alignment} or Eq.\ref{eq:kl_alignment_2} and weighted with a hyperparameter $\lambda$. Overall, $\mathcal{L}_{2^{\text{nd}}\;\text{stage}}$ maintains both clean and adversarial classification accuracy, while enforcing adversarial‑invariant cross‑modal alignment to bring adversarial embeddings into distributional agreement with clean embeddings across all classes.

\section{Experiment}
Our model configuration is inherited from LanguageBind. All modality encoders take ViT-L/14 as the backbone. The text encoder, which is an ViT-L/14 initialized from OpenCLIP, is locked while encoders of other modalities are fine-tuned through Low-Rank Adaptation(LoRA) during training. Based on the publicly released work of LanguageBind, the architectures for our multi-modal encoders are: the pretrained ViT-L/14 by OpenCLIP as the image encoder, fully fine-tuned ViT-L/14 as the audio and video encoders, and LoRA-tuned ViT-L/14 as the thermal encoder.

\noindent \textbf{Datasets.} For convenient comparison with LanguageBind, we train and evaluate on ImageNet-1k~\cite{5206848}, ESC-50~\cite{piczak2015dataset}, LLVIP~\cite{jia2021llvip}, and MSR-VTT~\cite{7780940}, corresponding to \textit{Image}, \textit{Audio}, \textit{Thermal}, and \textit{Video}. ImageNet-1k follows its official split, while the others are partitioned 8:2 per class. After stage1 fine-tuning, we evaluate on the full test sets. For stage2, limited batch size and fair evaluation considerations lead us to sample 10 examples per class (or all if fewer), ensuring sufficient coverage.

\noindent \textbf{Implementation Details.} All experiments are conducted on 8 vGPUs (32GB each). We train for 1 epoch per run, and each full set of experiments requires approximately 3 days of training time.

\noindent \textbf{Results.} Corresponding to the visualization in Figure~\ref{fig:radarmap}, Table~\ref{tab:rlbind_perf} numerically presents the performance of our method under APGD attacks of varying strength across 4 modalities - image, audio, thermal, and video - benchmarked against the baseline LanguageBind~\cite{zhu2024languagebind} and the RobustCLIP~\cite{schlarmann2024robustclip}. Although LanguageBind attains high clean accuracy, its robustness collapses under APGD attacks - for instance, \textit{image} accuracy drops from 74.07\% to 2.84\% and \textit{video} accuracy from 42.74\% to 0.13\% at $\epsilon=4/255$ — rendering it nearly unusable in adversarial settings. In stage1, RobustCLIP markedly boosts robustness — for instance, \textit{image} robustness rises from 9.12\% to 42.49\% at $\epsilon=2/255$ and from 2.84\% to 21.07\% at $\epsilon=4/255$, with \textit{thermal} gains exceeding 25\%. However, this comes at the cost of generalization, as clean accuracy in \textit{thermal} drops by up to 19.14\%. In Stage 2, \name{} leverages cross-modal alignment to markedly enhance robustness while restoring — and in some cases even surpassing — LanguageBind's clean accuracy. \textit{Image} robustness improves by up to 47.64\% with clean accuracy exceeding the baseline, \textit{thermal} maintains over 85\% across attacks, and \textit{audio} achieves a better balance than stage1, demonstrating \name’s ability to preserve generalization while substantially boosting robustness.

\subsection{Ablation Study}

\begin{table*}[h!]
\centering
\caption{Performance comparison under different fine-tuning strategies.}
\setlength{\tabcolsep}{18pt}
\scriptsize
\resizebox{\linewidth}{!}{%
\begin{tabular}{l c c c c}
\toprule
\textbf{Method} & \textbf{LoRA} &
\textbf{Clean Acc.} &
\makecell{\textbf{Robust Acc.}\\$\varepsilon_1=\tfrac{2}{255}$} &
\makecell{\textbf{Robust Acc.}\\$\varepsilon_2=\tfrac{4}{255}$} \\
\midrule

\multirow{2}{*}{LanguageBind + FARE2 (Stage1)}
& $\times$ & \textbf{77.88\%} & \textbf{55.46\%} & \textbf{31.93\%} \\
& \checkmark & 70.57\% & 42.49\% & 15.72\% \\
\hdashline

\multirow{2}{*}{RLBind2 (Stage2)}
& $\times$ & \textbf{83.95\%} & \textbf{66.33\%} & \textbf{37.36\%} \\
& \checkmark & 75.43\% & 56.76\% & 28.49\% \\
\midrule

\multirow{2}{*}{LanguageBind + FARE4 (Stage1)}
& $\times$ & \textbf{76.93\%} & \textbf{59.16\%} & \textbf{49.10\%} \\
& \checkmark & 75.98\% & 38.14\% & 21.07\% \\
\hdashline

\multirow{2}{*}{RLBind4 (Stage2)}
& $\times$ & \textbf{83.22\%} & \textbf{69.13\%} & \textbf{57.49\%} \\
& \checkmark & 79.08\% & 51.60\% & 33.23\% \\
\bottomrule
\end{tabular}%
}

\vspace{0.5em}
\raggedright
{\scriptsize
* RLBind-related experiments are with Dot-Product (Eq.\ref{eq:dot product}) as Cross-modal Correspondence and L2 (Eq.\ref{eq:kl_alignment_2}, $p=2$) as Adversarial-Invariant Cross-modal Alignment.\\
}
\label{tab:abl_lora}
\vspace{-12pt}
\end{table*}

To validate \name’s design, we conduct ablations on training stages, training objective, cross-modal correspondence(Eq.\ref{eq:dot product} to Eq.\ref{eq:mlp}) and alignment(Eq.\ref{eq:point-wise distribution alignment} or Eq.\ref{eq:kl_alignment_2}), and PEFT (LoRA). Results are shown in the following sections.

\subsubsection{Stages}

Table~\ref{tab:abl_stg} compares the performance of different stages in our design. We observe consistent accuracy gains of \name(stage2) over LanguageBind + FARE(stage1) across modalities.

\subsubsection{Adversarial-invariant Cross-modal Alignments}

Table~\ref{tab:abl_aca} provides a detailed comparison of how different configurations of cross-modal correspondence and alignment affect \name’s performance. In terms of cross-modal feature correspondence, dot product(Eq.~\ref{eq:dot product}) yields the best outcomes, followed by bilinear mapping(Eq.~\ref{eq:bilinear}) where a single neural layer is used to align features. Non-linear mapping(Eq.~\ref{eq:mlp}) performs worst, as the two-layer MLP tends to overfit given the current data scale. For adversarial-invariant alignment strategies, distribution level alignment(Eq.~\ref{eq:kl_alignment_2}) generally offers stronger robustness than point-wise alignment (Eq.~\ref{eq:kl_alignment_2}), albeit with a slight trade-off in clean accuracy. For a better robustness-generation balance, we choose dot-product(Eq.~\ref{eq:dot product}) as our cross-modal correspondence and point-wise distribution $L_{2}$(Eq.~\ref{eq:point-wise distribution alignment}, $p=2$) as our cross-modal alignment strategy.

\subsubsection{Loss Functions}

Table~\ref{tab:abl_lossfunc} provides a comparison of \name’s performance under different training objective configurations. Several trends emerge. \textbf{i)}, training with only clean cross-entropy achieves strong clean accuracy (e.g., 79.84\% for \name2), but robustness collapses, dropping to 26.66\% under $\epsilon = 2/255$ and 13.12\% under $\epsilon = 4/255$. This highlights the vulnerability of models optimized solely for clean performance. \textbf{ii)}, incorporating adversarial loss into the objective effectively mitigates this issue. For instance, \name2 with adversarial training shows only a modest decline in clean accuracy (79.84\% → 74.80\%), yet robustness more than doubles across both perturbation levels (26.66\% → 56.12\% at $\epsilon = 2/255$; 13.12\% → 27.57\% at $\epsilon = 4/255$). \textbf{iii)}, extending the objective with adversarial-invariant cross-modal alignment yields consistent improvements in both robustness and generalization. For example, \name4 with CMA maintains high clean accuracy (77.39\%) while achieving 51.62\% robustness at $\epsilon = 2/255$ and 33.02\% at $\epsilon = 4/255$, outperforming the non-aligned counterpart, especially under stronger attacks. Together, these findings confirm that adversarial-invariant cross-modal alignment plays a pivotal role in alleviating the robustness–accuracy trade-off and provides a reliable means of strengthening model resilience without sacrificing generalization.

\subsubsection{LoRA v.s. Fully Fine-tuning}

Table~\ref{tab:abl_lora} compares full fine-tuning and LoRA across different strategies. As expected, fully fine-tuned models achieve higher robustness and clean accuracy, e.g., RLBind2 reaches 83.95\% clean accuracy and 66.33\% robustness at $\epsilon=2/255$, clearly outperforming its LoRA counterpart. LoRA, however, offers competitive results with far fewer parameters and lower cost — RLBind4 with LoRA still attains 79.08\% clean accuracy and 33.23\% robustness at $\epsilon=4/255$. Overall, full fine-tuning delivers the strongest robustness, but LoRA provides a favorable efficiency–performance trade-off. Importantly, RLBind consistently improves robustness under both settings, underscoring its potential for scalable multimodal deployment.

\section{Conclusion}

We present \name, a two-stage adversarial-invariant cross-modal alignment framework designed to enhance the robustness of LanguageBind embeddings. To the best of our knowledge, we are the \textbf{first} to leverage cross-modal feature distribution alignment to protect LanguageBind’s embedding space from adversarial attacks, achieving robust performance without sacrificing generalization. Our extensive experiments demonstrate that \name{} consistently outperforms the LanguageBind backbone and baseline fine-tuning methods in both clean accuracy and norm-bounded robustness across \textit{Image}, \textit{Audio}, \textit{Thermal}, and \textit{Video} data. These results highlight the effectiveness of cross-modal feature alignment in strengthening multi-modal encoders, particularly for applications in robotics, where robust perception is crucial for tasks like navigation and manipulation.

Looking ahead, we aim to scale our approach to larger datasets, such as VIDAL-10M, and explore advanced strategies, including adaptive anchors and distribution-aware objectives, to further enhance both robustness and generalization. By addressing the critical vulnerabilities in multi-modal models, \name{} paves the way for safer deployment of such models in real-world robotic systems.

\bibliography{IEEEexample}
\bibliographystyle{IEEEtran}


\end{document}